# Applying an Ensemble Learning Method for Improving Multi-label Classification Performance

Amirreza Mahdavi-Shahri, Mahboobeh Houshmand, Mahdi Yaghoobi, Mehrdad Jalali
Department of Computer Engineering,
Mashhad Branch, Islamic Azad University
Mashhad, Iran
{amirrezamahdavi-sh, houshmand, yaghoobi}@mshdiau.ac.ir, mehrdadjalali@ieee.org

*Abstract*—in recent years, multi-label classification problem has become a controversial issue. In this kind of classification, each sample is associated with a set of class labels. Ensemble approaches are supervised learning algorithms in which an operator takes a number of learning algorithms, namely base-level algorithms and combines their outcomes to make an estimation. The simplest form of ensemble learning is to train the base-level algorithms on random subsets of data and then let them vote for the most popular classifications or average the predictions of the base-level algorithms. In this study, an ensemble learning method is proposed for improving multi-label classification evaluation criteria. We have compared our method with well-known base-level algorithms on some data sets. Experiment results show the proposed approach outperforms the base well-known classifiers for the multi-label classification problem.

## I. INTRODUCTION

In a conventional single-label classification [1-2], training samples are associated with a single label *l* from an already known finite set of disjoint labels L. In this method, a single label dataset $D$ consists of $n$ samples, $(m_1, l_1), (m_2, l_2), \ldots, (m_n, l_n)$ that $m$ represents the input data (consist of some attributes) and *l* represents the single label to which *m* belongs. If the number of labels is equal to two, then the learning task is referred to as binary classification and if it is more than two, it is called multi-label classification. Multi-label classification is conceptually different with multi-class classification. In multi-class classification, the classifying samples map into just one of the more than two classes, while multi-label classification also allows samples to belong to multi classes.

In contrast with the single-label learners, the multi-label method is impacted by intrinsic latent correlations between all labels, which indicates that the membership of a sample instance to a class can help to estimate its set of labels. For example, a patient with a high blood pressure is more likely to suffer a heart disease than a person with a normal blood pressure, but less likely to develop a neuromuscular diseases [3-4].

Multi-label classification method have been applying in applications such as bioinformatics where each protein may be labeled by multiple functional labels like metabolism, energy or such in cellular biogenesis [3-4], or in video annotation, a movie can be defined with some labels or some tags [5]. Another application of multi-label classification is in categorization of texts, where each document can be assigned to a set of predefined topics [6-7].

Basically, multi-label classification approaches can be categorized in two different groups: I) problem transformation methods. The problem transformation methods intend to transform multi-label classification tasks into one or more single-label classification problems [8-10], [14], and II) algorithm adaptation methods. The algorithm adaptation methods extend traditional classifiers to handle multi-label problems directly [11-14]. Moreover, there exist multi-label extensions of support vector machine [15], neural network [16] and decision tree [17] or other learning algorithms.

The most common problem transformation method considers the prediction of each label as an independent binary classification task. It learns one binary classifier $h_\lambda : X \to \{\neg\lambda, \lambda\}$ for every different label $\lambda \in L$. It then transforms the original data set into |L| data sets $D_\lambda$ such that it contains all examples of the principal data set, labeled as $\lambda$ if the labels of the original example contains $\lambda$ and as $\neg\lambda$ otherwise. The same approach can be used in order to deal with a multi-class problem using a binary classifier, typically referred to as one-against-all or one-versus-rest. Following [14], this method is called Binary Relevance (BR) learning. BR is criticized for not considering correlations between the labels [1].

Several problem transformation methods exist for multi-label classification where the base-line approach is called the binary relevance method. The random *k*-label sets (RAKEL) algorithm uses multiple lower power set (LP) classifiers, each trained on a random subset of the actual labels [18]. A main challenge in the multi-label learning systems is that the number of feasible label combination increases exponentially. Conventional multi-label learning methods insist on utilizing the correlations between labels to improve the accuracy of individual multi-label learner [19-22].

Ensemble is a supervised learning algorithm in which an agent takes a number of classifiers and then combines their outputs to make a prediction. The classifiers being combined are called base-level ones.

In recent years, ensemble techniques play an important role in the research field of data mining and machine learning, because they can improve accuracy of classifiers individual classifiers. [23-24]. Ensembles methods are well-known for overcoming over-fitting problems which decrease the generalization of systems, particularly in highly unbalanced data sets [23]. These approaches are either homogeneous or heterogeneous. For the first state, base-level classifiers are constructed using the same algorithms, and in the other state, base-level classifiers are constructed using various algorithms to improve performance.

The utilization of ensemble learning is widespread in the last decade, for example in [19] base classifiers are used as the landscape of ensemble classifiers, or in [20], the problem of multi-label selective ensemble is studied.

The main idea of this study is using a heterogeneous ensemble of multi-label learners in order to get better results. Another benefit of combining classifiers which are multi-label is that both correlation and imbalance problems can be tackled together. Ensemble of multi-label classifiers can handle the imbalance problem while the correlation can be handled by using the state-of-the-art multi-label classifiers [25] [26] as base classifiers that inherently consider the correlation among labels.

The proposed ensemble learning (EN-MLC) is applied to three publicly available multi-label data sets from different domains, namely Scene, Yeast, and Music and five different multi-label classification metrics are computed and compared with the ones for five different base-algorithm classifiers.

The rest of this paper is organized as follows. The background material is explained in Section II. Section III presents the proposed method. In Section IV the results are explained and finally Section V concludes the paper.

## II. BACKGROUND

The traditional multi-label learning methods focus on finding the correlations between labels in order to increase the accuracy of individual multi-label learners [18-20]. Based on the strategy of constructing of the learning system, these approaches can be typically classified into the following two categories (1) Multi-label learning methods based on a group of single-label learners (Figure 1(a)), for example EPS or RAKEL [1] and (2) Multi-label learning methods based on individual multi-label learners (Figure 1(b)). In these kinds of methods, a multi-label learner is established in order to make estimation on labels. The correlations between all labels are utilized in the models or learning systems of the multi-label learner, e.g., the neural network structure in ML-RBF [28].

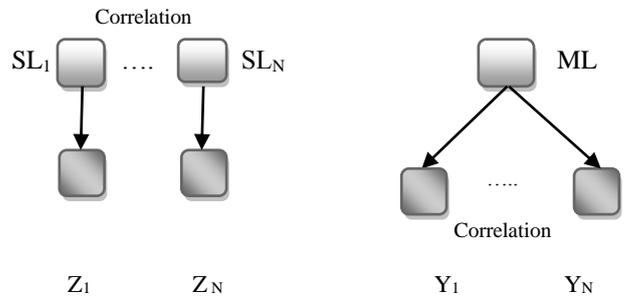

(a) Single-label learners  (b) Individual multi-label learner

Fig.1. modeling multi-label learning system. *SL* and *ML* represent the single-label and multi-label learner, respectively. *Y* and *Z* show the single and atomic label, respectively [27].

### A. Evaluation Metrics

Single-label classification is different with multi-label classification. In single-label classification, training samples can be correct or incorrect, but in multi-label classification samples are either limitedly correct or incorrect. This can happen when a classifier correctly assigns an example to at least one of the labels it belongs to, but not to the all labels it belongs to. In addition, a classifier may also assign to an example to one or more labels to which it does not belong [21]. In this sense, the evaluation of multi-label classifiers needs different tools than those used in single-label methods. Some approaches have been proposed in the literature for the assessment of multi-label classifiers.

In [2], the measures can be widely categorized in two classes: bipartition-based and ranking-based. Some of the bipartition-based measures, namely example-based-measures, evaluate bipartition over all examples of the evaluation data set. Moreover, the ranking-based measures evaluate rankings with respect to the ground truth of multi-label data set.

In this paper, five measures are selected for the comparison of the proposed method with formerly existing multi label classification approaches, which are introduced in the following.

An evaluation dataset can be defined as $(x_i, Y_i)$, $i=1\ldots N$, where $Y_i \subseteq L$ is the set of true labels and $L = \{\lambda_j : j = 1,\ldots,M\}$ is the set of all labels. Given an example $x_i$, the set of labels which are estimated by a multi-label approach is shown by $Z_i$, while the rank predicted for a label $\lambda$ is denoted as $r_i(\lambda)$. The most pertinent label gets the highest rank (1), while the least pertinent one receives the lowest rank (*M*) [2].

*Bipartition-based Measures:*

*Accuracy:*

Accuracy computes the sum of correct labels divided by the union of predicted and true labels as defined in (1).

$$Accuracy = \frac{1}{N}\sum_{i=1}^{N}\frac{|Y_i \cap Z_i|}{|Y_i \cup Z_i|} \qquad (1)$$

*Hamming Loss:*

Hamming Loss (as defined in (2)) considers estimation errors that are called incorrect label and missing errors. It evaluates the frequency where an example-label pair is misclassified, i.e., an example is assigned to an incorrect label or a label belonging to the instance is not properly predicted. When the value of hamming loss is decreased, the better the performance is obtained and the best case is occurred when it is equal to 0.

$$Hamming\ Loss = \frac{1}{N}\sum_{i=1}^{N}\frac{|Y_i \Delta Z_i|}{M} \quad (2)$$

*Ranking-based Measures:*

*One-error:*

One-error (3) measure estimates the number of times the top-ranked label was not in the set of possible labels. For single label classification problems, the one-error is similar to ordinary error. The better performance is achieved for the smaller value of one-error.

$$One-error = \frac{1}{N}\sum_{i=1}^{N}\delta\left(\arg\min r_i(\lambda)\right) \quad (3)$$

Where:

$$\delta(\lambda) = \begin{cases} 1 & if\ \lambda \in Y_i \\ 0 & otherwise \end{cases}$$

*Ranking Loss:*
Ranking loss considers the frequencies of incorrect outcome values as defined in (4).

$$Ranking\ loss = \frac{1}{N}\sum_{i=1}^{N}\frac{1}{|Z_i||\overline{Z_i}|} \quad (4)$$

*Average Precision:*
Average precision, as computed in (5), which is an overall measure to evaluate an algorithm, is the average precision for all the possible labels. It measures the average fraction of labels ranked superior a particular label $L \in Y_i$ which is actually in $Y_i$. The better performance is obtained for the bigger value of average prevision and the best case is when it is equal to 1.

$$avgprec(h;U) = \frac{1}{N}\sum_{i=1}^{N}\frac{1}{|Y_i|}\sum_{y \in Y_i}\frac{|P_i|}{rank^f(X_i, y)} \quad (5)$$

Where:

$$P_i = \{y' \mid rank^f(x_i, y), y' \in Y_i\}$$

## III. THE PROPOSED METHOD

Ensemble learning which a supervised learning is a solution to the given problem. Suppose *X* denote a set of instances and let $Y = \{1, 2, ..., N\}$ be a set of labels. Given a training set $S = \{(x_1, y_1), ...., (x_m, y_m)\}$ that $x_i \in X$ is a single case and $y_i \subseteq Y$ is the label set assigned with $x_i$, we intend to design a multi-label classifier *H which* predicts a subset of labels for an unseen sample. Ensemble of multi-label classifiers trains *q* multi-label classifiers, namely $H_1$, $H_2$... $H_q$. Therefore, q patterns are diverse and able to give various multi-label predictions. For an unseen sample $x_j$, each $k^{th}$ individual pattern (from *q* patterns) give an *N*- dimensional vector $P_{jk} = [p_{1k}, p_{2k}, ....., p_{Nk}]$, where the value $p_{bk}$ is the probability of the $b^{th}$ class label associated with classifier *k* being correct. Figure 2 shows the strategy of constructing multi-label ensemble learning systems.

We apply an ensemble learning method to make a group of single-label base-learners. The base-learners in the ensemble are to make a prediction on a single label. Then, those base-learners are combined as one multi-label learner to make predictions on all labels. Correlations among labels are utilized between these single-label learners.

First the model deals with LP complexity and prunes samples with rare label combinations to let the model focus on the most important label sets. Then, it compensates the information loss by reintroducing the pruned sample associated with subset of their original label set. It is noteworthy that an LP model is not able to output label sets that are not in the training set. To tackle this problem, ensemble multi-label classification, EN-MLC combines the results of several classifiers where each base classifier is an LP trained on a random selection of samples.

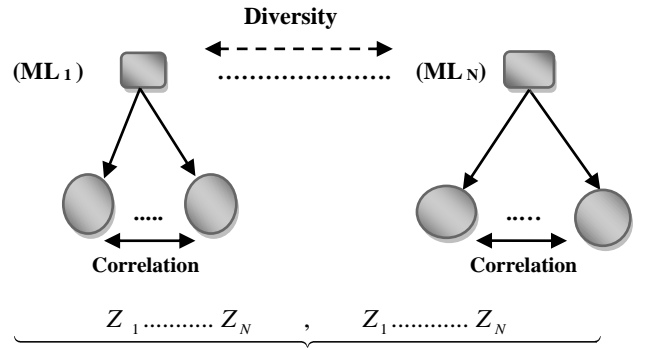

Fig. 2. Strategy of multi-label ensemble learning system, Z represents the single label.

To combine outputs of these classifiers, there are some methods such as an average, weighted average, maximum and minimum that are called algebraic methods and also majority voting, weighted majority voting are called voting methods. Ensemble learning is used to construct a group of single-label base learners. Base-learners consist of structures to estimations on a single label and finally those base-learners are combined to make predictions on all labels. In this paper, we apply the majority voting approach for this combination.

In the following the method is described. A group of accurate and diverse multi-label base-learners are considered. The main difference between former ensemble methods for multi-label leaning algorithms is that base-learners in the multi-label ensemble learning are not single-label learners, but they are multi-label learners. Algorithm 1 shows the pseudo-code of the proposed approach.

---

**Algorithm 1  EN-MLC**

**Input:** A: new instance $x$ ($k$-labelset matrix – $M$)
H consists of base classifiers
Count: # of base classifiers
$T_1$ and $T_2$ are train & test sets,
**Output:** $Y(j) \ \forall \ \# \ of$ ensemble of Multi classifiers $g$

**Procedure TRAINING:**
1. **While (1)**
// $Y_i$: the labelset represented by the $i$-th row in $M$;
Repeat steps 2-5
// EN-MLC
2. **For** $i=1$ to the number of rows in $M$
$U_i$ = increasing value metrics
$L_i$ = decreasing value metrics
3. $H_t = \{ \# \ of \ U_t \cup \# \ of \ L_t \}$
4. **If** $(Min \ L_t > Min_{ENML}) \wedge (Max \ U_t < Max_{ENML})$
5. $D_{i,j} \leftarrow (Min_{EN-MLC}, Max_{EN-MLC})$
end for
end procedure

---

## IV. EXPERIMENTAL RESULTS

### A. The Used Dataset

Three multi-label data sets, namely Scene, Yeast, and Music, selected from different domains, as introduced in the following are used.

Scene: Image data set scene [29] contains 2407 images associated with up to six signification and the number of labels, like beach, mountain and field is 6.

Yeast: The data set yeast [15] is related to protein function classification. It contains micro-array expressions and phylogenetic profiles that have 2417 yeast genes. Every gene is presented by a subset of 14 (number of labels) functional batches (e.g. Metabolism, energy, etc.)

Music: This data set consists of 592 samples and six labels.

Two parameters for the used data sets, namely, label cardinality (LCard) and label density (LDen) are considered which are defined in the following.

Definition 1: The label cardinality (LCard) from $D$ is the average number of labels per instance. It is computed as (6) where $N$ is the total number of instances:

$$LCard(D) = \frac{1}{N} \sum_{i=1}^{N} |Y_i| \quad (6)$$

Definition 2: Label density is number of labels in the classification approaches. Two data sets with the same label cardinality but with a considerable difference in the number of labels might not determine the same attributes that leads to different behaviors with respect to the multi-label classification approaches. Label density is defined as (7):

$$LDen(D) = \frac{1}{N} \sum_{i=1}^{N} \frac{|Y_i|}{|L|} \quad (7)$$

Where $L = \bigcup_{i=1}^{N} Y_i$.

In Table 1, a summary of the used data set statistics is provided.

TABLE I.   THE USED DATA SET STATISTICS

| Dataset | #instances | #train | #test | #labels | weights | LCard | LDen |
|---|---|---|---|---|---|---|---|
| *Scene* | 2407 | 1588 | 819 | 6 | 2407 | 1.08 | 0.18 |
| *Yeast* | 2417 | 1595 | 822 | 14 | 2417 | 4.23 | 0.302 |
| *Music* | 592 | 390 | 202 | 6 | 592 | 1.827 | 0.24 |

### B. Setup of the Algorithms

There are some frameworks for doing a multi-label classification task such as sickie-learn [30], Orange [31], but they include only few number of multi-label classification algorithms. All implementations of this study are performed in Weka-based package of Java's classes that is called Mulan[1]. This package includes multiple methods of classifications like LP and IRAKEL and ensemble of classifiers. Mulan is composed of different libraries for performing various classifiers.

### C. Performance Analysis

We used five different base-level classifiers and compared outputs with them. The classifiers are K-NN, Naïve Bayes (NB), RANDOM TREE (RT), REPTREE and J48. We ran several tests to estimate the best possible performance for each classifier and chose it as the output of that classifier.
Tables II, III, and IV show the outputs of applying these classifiers to the Scene, the Yeast and the Music the data sets respectively. These classifiers are learnt by the state-of the art improved BR (IBR) [1] and improved RAKEL (IRAKEL) learning algorithms [21].

The best values for each metric in applying classifiers are bolded. The average of these metrics for all classifiers are shown in the last columns of tables.

---

[1]https://sourceforge.net/projects/mulan/

TABLE II. RESULTS OF APPLYING FIVE CLASSIFIERS ON THE SCENE DATA SET

| IRAKEL | NB | k-NN | RANDOM-T | REPTREE | J48 | AVERGE |
|---|---|---|---|---|---|---|
| Acc ↑ | 0.52 | **0.67** | 0.571 | 0.618 | 0.614 | 0.598 |
| HL↓ | 0.151 | **0.113** | 0.151 | 0.114 | 0.131 | 0.132 |
| 1-Err ↓ | 0.377 | 0.302 | 0.325 | **0.289** | 0.293 | 0.317 |
| RL ↓ | 0.162 | 0.184 | 0.154 | **0.137** | 0.142 | 0.155 |
| AvPre ↑ | 0.322 | **0.335** | 0.325 | 0.322 | 0.334 | 0.327 |
| IBR | NB | k-NN | RANDOM-T | REPTREE | J48 | AVERAGE |
| Acc ↑ | 0.457 | 0.182 | 0.5 | 0.376 | **0.506** | 0.404 |
| HL ↓ | 0.178 | 0.818 | 0.146 | 0.256 | **0.143** | 0.308 |
| 1-Err ↓ | 0.397 | **0.297** | 0.437 | 0.419 | 0.422 | 0.394 |
| RL ↓ | **0.146** | 0.185 | 0.228 | 0.17 | 0.278 | 0.201 |
| AvPre ↑ | —— | 0.337 | 0.351 | **0.405** | 0.375 | 0.367 |

TABLE III. RESULTS OF APPLYING FIVE CLASSIFIERS ON THE YEAST DATA SET

| IRAKEL | NB | k-NN | RANDOM-T | REPTREE | J48 | AVERAGE |
|---|---|---|---|---|---|---|
| Acc ↑ | 0.419 | 0.493 | 0.407 | **0.498** | 0.417 | 0.446 |
| HL ↓ | 0.324 | **0.239** | 0.336 | 0.244 | 0.324 | 0.293 |
| 1-Err ↓ | 0.44 | 0.427 | 0.437 | **0.361** | 0.422 | 0.417 |
| RL↓ | 0.304 | 0.269 | 0.302 | **0.239** | 0.285 | 0.279 |
| AvPre ↑ | 0.379 | 0.393 | **0.396** | 0.383 | 0.382 | 0.386 |
| IBR | NB | k-NN | RANDOM-T | REPTREE | J48 | AVERAGE |
| Acc ↑ | 0.404 | **0.493** | 0.388 | 0.467 | 0.435 | 0.437 |
| HL ↓ | 0.281 | 0.239 | 0.278 | **0.238** | 0.256 | 0.258 |
| 1-Err↓ | 0.358 | 0.448 | 0.499 | **0.335** | 0.481 | 0.424 |
| RL↓ | 0.259 | 0.269 | 0.309 | **0.21** | 0.313 | 0.272 |
| AvPre ↑ | **0.406** | 0.394 | 0.387 | 0.373 | 0.313 | 0.374 |

TABLE IV. RESULTS OF APPLYING FIVE CLASSIFIERS ON THE MUSIC DATA SET

| IRAKEL | NB | k-NN | RANDOM-T | REPTREE | J48 | AVERAGE |
|---|---|---|---|---|---|---|
| Acc ↑ | 0.455 | 0.49 | 0.477 | **0.526** | 0.523 | 0.494 |
| HL↓ | 0.276 | 0.248 | 0.296 | **0.241** | 0.246 | 0.261 |
| 1-Err↓ | 0.361 | 0.401 | 0.361 | 0.302 | **0.297** | 0.344 |
| RL↓ | 0.23 | 0.294 | 0.234 | 0.221 | **0.185** | 0.232 |
| AvPre ↑ | 0.422 | **0.442** | 0.421 | 0.417 | 0.411 | 0.422 |
| IBR | NB | k-NN | RANDOM-T | REPTREE | J48 | AVERAGE |
| Acc ↑ | 0.503 | 0.304 | 0.443 | **0.508** | 0.39 | 0.429 |
| HL ↓ | **0.231** | 0.696 | 0.26 | 0.233 | 0.318 | 0.347 |
| 1-Err ↓ | 0.337 | 0.396 | 0.48 | 0.356 | 0.49 | 0.411 |
| RL↓ | **0.199** | 0.291 | 0.303 | 0.226 | 0.295 | 0.262 |
| AvPre ↑ | **0.505** | 0.448 | 0.443 | 0.502 | 0.455 | 0.470 |

Then, we applied the proposed EN-MLC method and compared its results with the average metric with respect to the five evaluation metrics.

Figures 2, 3 and 4 indicate that the proposed approach leads to better results as compared to the average of base-level classifiers with respect to four metrics, namely accuracy (Acc), Hamming loss (HL), one-error (1-Err), and Ranking loss (RL) on the Scene, the Yeast and the Music datasets, respectively.

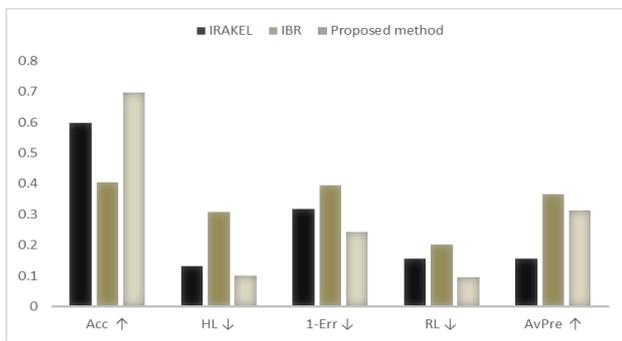

Fig 2. Results of the proposed method comparison with IRAKEL [21] and IBR [1] algorithms on the Scene data set

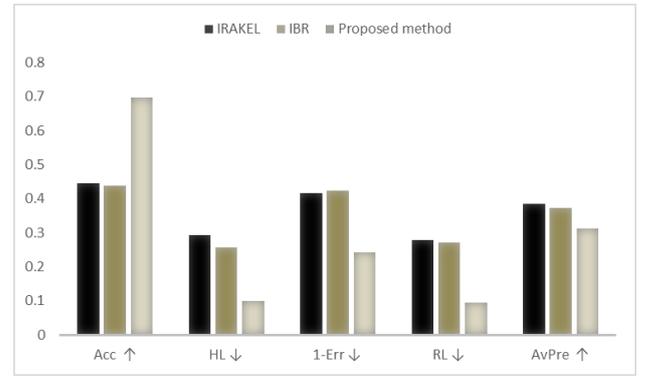

Fig 3. Results of the proposed method comparison with IRAKEL [21] and IBR [1] algorithms on the Yeast data set

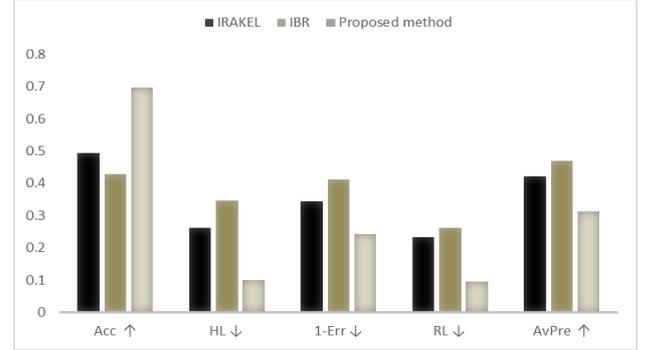

Fig 4. Results of the proposed method comparison with IRAKEL [21] and IBR [1] algorithms on the Music data set

## V. CONCLUSION AND FUTURE WORKS

In this paper, application of ensemble learning for improving multi-label classification was considered. Data sets from various domains were selected and base-level classifiers and the proposed ensemble learning were applied to them. Five known metrics in the multi-label classification domain were computed for these approaches. Experimental results showed the proposed ensemble method (EN-MLC) lead to better results than the base-level algorithms, with respect to the four out of five evaluation metrics. As a future work, applying other classification approaches such as the ones based on back-propagation learning algorithm and semi-supervised learning approaches is being considered.

*ACKNOWLEDGMENT*

We would like to express our gratitude to Prof. Mohammad-R. Akbarzadeh-T and Dr. Mojtaba Asgari for their helps during preparing this study.